\documentclass[conference]{IEEEtran}
\IEEEoverridecommandlockouts
\usepackage{cite}
\usepackage{amsmath,amssymb,amsfonts}
\usepackage{algorithmic}
\usepackage{graphicx}
\usepackage{textcomp}
\usepackage{url}
\usepackage{xcolor}
\usepackage{fancyhdr}
\fancypagestyle{IEEEtitlepagestyle}{
  \fancyhf{}
  \fancyhead[L]{\small IEEE NIGERCON 2026}
  \fancyfoot[C]{\footnotesize 979-8-3195-3455-2/26/\$31.00~\copyright 2026 IEEE}

}
\fancypagestyle{fancy}{
  \fancyhf{}
  \fancyhead[L]{\small IEEE NIGERCON 2026}

}
\def\BibTeX{{\rm B\kern-.05em{\sc i\kern-.025em b}\kern-.08em
    T\kern-.1667em\lower.7ex\hbox{E}\kern-.125emX}}
    
\begin{document}

\title{Adaptive Low-Light Enhancement for Road Anomaly Detection: A
Multi-Architecture Benchmark and Edge Deployment Study
}

\author{\IEEEauthorblockN{ Obed Honour Eje}
\IEEEauthorblockA{\textit{Mechatronics Engineering} \\
\textit{Federal University of Technology Minna}\\
Minna, Nigeria \\
obed.m2200396@st.futminna.edu.ng
}
\and
\IEEEauthorblockN{Oluwatosin Kolade}
\IEEEauthorblockA{\textit{Mechanical Engineering} \\
\textit{Obafemi Awolowo University}\\
Ile-Ife, Nigeria \\
okolade@student.oauife.edu.ng}
\and
\IEEEauthorblockN{Glory Bagai}
\IEEEauthorblockA{\textit{Mechatronics Engineering} \\
\textit{Federal University of Technology Minna}\\
Minna, Nigeria \\
kuzayet.m2204354@st.futminna.edu.ng}
\and
\IEEEauthorblockN{Taliha Abiodun Folorunso}
\IEEEauthorblockA{\textit{Mechatronics Engineering} \\
\textit{Federal University of Technology Minna}\\
Minna, Nigeria\\
funso.taliha@futminna.edu.ng}
}

\maketitle
\thispagestyle{IEEEtitlepagestyle}

\begin{abstract}
Road anomaly detection systems on autonomous vehicles must function reliably under degraded visual conditions including darkness, rain, and Harmattan dust haze. This paper benchmarks four object detection architectures, YOLOv8n, YOLOv12n, RT-DETR, and Faster R-CNN, under physically grounded lighting and weather degradation, and shows that uniform application of the SCI low-light enhancement network is actively harmful at dusk-level lighting ($-16.1\%$ to $-30.7\%$ mAP@50). We propose a lightweight, training-free luminance-adaptive router that selects between no enhancement, moderate SCI, and denoising plus strong SCI based on per-frame Rec. 601 luminance, recovering most of the dusk penalty while preserving low-light gains, and we validate the full pipeline on NVIDIA Jetson Nano edge hardware.

\end{abstract}


\section{Introduction}
Road-traffic crashes remain a severe public-health and economic burden in Nigeria: the Federal Road Safety Corps recorded 5,421 deaths from 9,570 crashes in 2024, a 7\% rise over 2023 [1], and the WHO places Nigeria's road-traffic death rate at 21.4 per 100,000, among the highest in Africa [2]. Poor infrastructure, potholes, eroded shoulders, poor lighting, weak signage, is repeatedly implicated, and the World Bank estimates a socio-economic cost of USD 29 billion in 2016, over 7\% of GDP [3]. Automated pothole and speed-bump detection is a safety-critical perception task for advanced driver-assistance systems (ADAS) and autonomous vehicles on West African roads, yet most detectors are trained and validated under favourable daytime illumination. Reliability degrades sharply under conditions routine in Nigeria: unlit night-time roads, the rainy season, and the Harmattan, a seasonal Saharan dust haze that can cut visibility to 200–500 m between November and February [6]. These conditions reduce contrast, add signal-dependent noise, and increase atmospheric scattering, weakening the features detectors depend on.

A natural mitigation is a front-end enhancement module that restores visibility before detection [9], but enhancement optimised for human perception does not necessarily benefit, and can actively harm, machine vision, since the two objectives are misaligned [10], [11]. Retraining the detector on augmented data is more robust but computationally expensive and impractical to repeat per degradation type on resource-constrained edge hardware such as the Jetson Nano.

Vision-based road-damage detection has moved from classical image processing toward CNN detectors, with YOLO variants now predominant [5], [15]; multi-country benchmarks such as RDD2020/2022 [16, 17] target daytime conditions, few studies address low light explicitly [18], and African contexts remain under-represented. Detector design spans two-stage detectors such as Faster R-CNN [14], one-stage detectors such as the YOLO series [4], [12], [19], and real-time transformer detectors such as RT-DETR, the first DETR-style model to outperform comparable YOLOs in speed and accuracy [13]; comparative benchmarking of these families under degradation remains sparse. Low-light enhancement has advanced from Retinex theory to deep networks such as Zero-DCE [7], EnlightenGAN [8], and the lightweight SCI network [9], but such enhancement can harm downstream detection through noise, edge blur, and colour artefacts [10], [11]. Adverse weather poses a related challenge: rain and haze are commonly modelled via the Koschmieder scattering equation [20], with physically grounded studies reporting detection drops of 25–30\% under extreme rain [22]. Harmattan haze is well characterised in the atmospheric-science literature [6] but is essentially absent from the detection-robustness literature, a gap this paper begins to address.

Closest to our approach, IA-YOLO [23], GDIP-YOLO [24], and DENet/DE-YOLO [25] integrate learned, always-on enhancement end-to-end with the detector, while YOLO-DER [26] and ClassLIE [27] make a binary enhance/bypass decision after additional training; none provides a training-free, three-way routing decision with a full-bypass branch for road-anomaly detection. Benchmarks confirm lightweight YOLO variants are typically the only models approaching real-time on a Jetson Nano [28], often requiring TensorRT, quantisation, or pruning [29]; we extend these benchmarks by jointly accounting for the added cost of the enhancement front-end. 

This paper investigates whether front-end image enhancement can serve as a practical, real-time perception-recovery strategy for road-anomaly detection on edge-deployed vehicles. Our contributions are fourfold: a four-architecture benchmark (YOLOv8n, YOLOv12n [4],[12], RT-DETR [13], Faster R-CNN [14]) under simulated lighting/weather degradation; evidence that uniform SCI enhancement harms dusk-level detection ($-16.1\%$ to $-30.7\%$ mAP@50); a lightweight, training-free luminance-adaptive router that recovers most of this loss; and a Jetson Nano edge-latency benchmark of the full pipeline.

\section{METHODOLOGY}

The overall architecture of the proposed perception-recovery pipeline is illustrated in fig. 1. Rather than forcing all images through a static enhancement front-end, the system ingests degraded frames and computes a lightweight perceptual luminance score to dynamically route the signal. Depending on this score, the frame is either bypassed entirely, passed through a moderate enhancement network, or heavily denoised and maximally enhanced before detection. This dynamic routing ensures that computational resources on constrained edge hardware are only expended when genuinely necessary to recover features.

\begin{figure}[t]
\centerline{\includegraphics[width=\columnwidth]{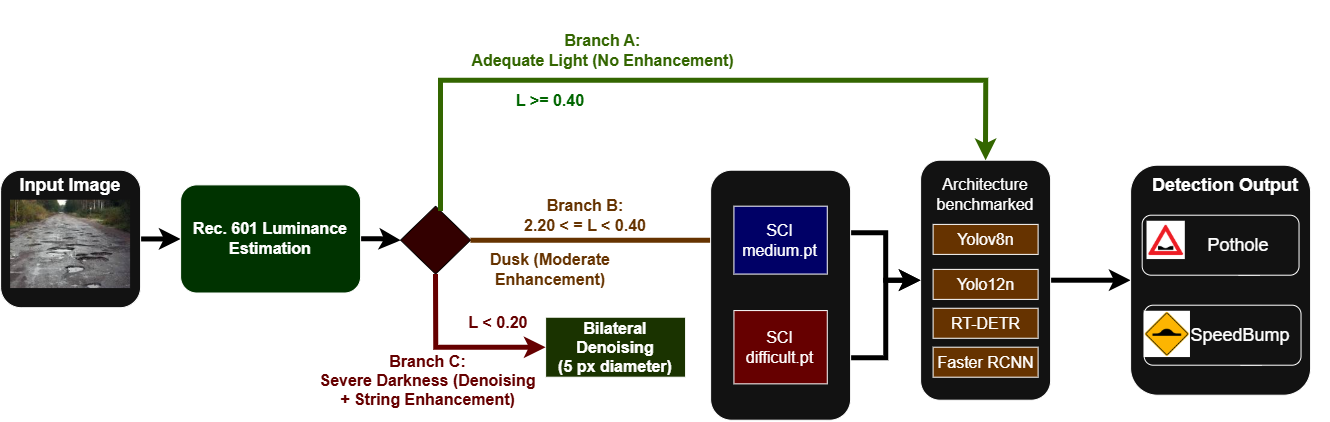}}
\caption{The proposed luminance-adaptive routing architecture.}
\label{fig:architecture}
\end{figure}

\subsection{Dataset Construction}
The evaluation corpus was constructed by merging two publicly available road-anomaly datasets under a targeted class-consolidation protocol: the Rajdalsaniya Pothole Detection Dataset (Kaggle, 2022; 2,077 images, single “pothole” class) [30] and the speed-unmarked-bump dataset v3 (Roboflow Universe, 2023; 2,367 images, “pothole” and “speed bump” classes, with manhole-cover variants excluded) [31]. The merged corpus comprises 4,444 images across two classes (pothole, speed bump), resized to 640 × 640 px and partitioned 70/15/15 into training, validation, and test splits. All degradation pipeline development and router threshold derivation were conducted on the 667-image validation split. The held-out 666-image test split was reserved for final confirmation of generalisability and evaluated after all methodological decisions were fixed.
\subsection{Degradation Pipeline}
A physically grounded degradation pipeline was built from scratch after 
preliminary trials with albumentations-based brightness-shift augmentation 
caused catastrophic model collapse (YOLOv8n map@50 at dusk fell 
from 0.88 to 0.056) due to unrealistic linear brightness artefacts. 
Lighting degradation was simulated via gamma compression combined with 
signal-dependent (Poisson-like) noise at four severity levels, calibrated to 
target mean luminance values: clean ($\gamma = 1.0$, no noise, mean 
luminance $\approx 0.462$), dusk ($\gamma = 1.6$, $\sigma = 0.025$, 
$\approx 0.332$), night ($\gamma = 2.8$, $\sigma = 0.060$, $\approx 0.207$), 
and severe ($\gamma = 4.2$, $\sigma = 0.110$, plus 5-pixel horizontal motion 
blur, $\approx 0.141$), with noise variance scaling with local pixel 
brightness to approximate Poisson shot noise. Weather degradation was 
simulated independently for three conditions: light rain (400 streaks, 7~px) 
and heavy rain (700 streaks, 10~px) via a custom scene-adaptive streak 
renderer, and Harmattan fog via the Koschmieder atmospheric-scattering 
equation [20] with a depth-gradient transmission map 
and a warm dust tint (RGB multiplier $[1.0, 0.97, 0.92]$). Lighting and 
weather degradations were applied independently, not compounded, and a 
visual sanity check preceded full-batch generation for each condition.

\subsection{Detection Architectures and Training}
Four detection architectures spanning three design paradigms were trained and evaluated under identical data splits: YOLOv8n [4] and YOLOv12n [12] (anchor-based and anchor-free, attention-enhanced one-stage CNN detectors, nano variants), RT-DETR [13] (transformer-based, end-to-end, anchor-free), and Faster R-CNN with a ResNet-50 Feature Pyramid Network backbone [14] (two-stage, region-proposal-based). YOLOv8n, YOLOv12n, and RT-DETR were trained using the Ultralytics framework; Faster R-CNN was trained using torchvision. All four models were trained on the clean (non-degraded) training split and evaluated, without retraining, on the degraded validation images, isolating the effect of test-time visual degradation and preprocessing from any domain-adaptation effect of training-time augmentation.

\subsection{Low-Light Enhancement Module}
Self-Calibrated Illumination (SCI) [9] was selected as the enhancement module for its favourable accuracy-to-cost ratio: 0.0003 M parameters, 0.0619 GFLOPs, and approximately 1.7 ms per frame on a TITAN X. The enhanced output R is obtained as follows:

\begin{equation}
L = \mathcal{H}_{\theta}(I), \quad \hat{R} = \text{clip}\left(\frac{I}{L},\ 0,\ 1\right)
\label{eq:sci}
\end{equation}

where I is the degraded input image and L is the illumination map estimated by the SCI network H .Two pretrained weight variants released by the original authors were used: medium.pt for moderate low-light conditions and difficult.pt for severely dark scenes. SCI was applied as a fixed, non-fine-tuned front-end preprocessor; it was not retrained or adapted to the road-anomaly domain. SCI was deliberately not evaluated on weather-degraded images: Harmattan fog frames are already high-luminance due to the white haze overlay, and applying SCI to them produced complete whiteout; weather-specific restoration (dehazing, deraining) was judged to require domain-specific modules and is left as future work.

\subsection{Luminance-Adaptive Router}
To address the finding that uniform SCI enhancement is harmful when the input signal is already adequate, a luminance-adaptive router was developed. Perceptual luminance is computed per frame using Rec. 601 weights, adding approximately 1.2 ms of overhead on a Jetson Nano:

\begin{equation}
L(I) = 0.299R + 0.587G + 0.114B
\label{eq:luminance}
\end{equation}

Routing thresholds were set as coarse round-number boundaries 
between the measured mean luminance centroids of each degradation 
level (dusk $\approx 0.332$, night $\approx 0.207$, severe $\approx 0.141$): 0.40 separates 
adequate-light from dusk, and 0.20 separates dusk from severe 
darkness. No threshold search was performed against detection 
metrics; the boundaries require only luminance statistics, not 
detection labels, preserving the router's training-free and 
label-free character.

\begin{equation}
\mathcal{P}(I) = \begin{cases}
I & \text{if } L \geq 0.40 \quad \text{(Branch A)}\\
\hat{R}_{\text{medium}}(I) & \text{if } 0.20 \leq L < 0.40 \quad \text{(Branch B)}\\
\hat{R}_{\text{difficult}}(\mathcal{D}(I)) & \text{if } L < 0.20 \quad \text{(Branch C)}
\end{cases}
\label{eq:router}
\end{equation}

The router is a fixed, rule-baseddecision function; it involves no learned parameters and required no additional training.

\subsection{Evaluation Protocol}
Detection accuracy was measured using mAP@50 (mean Average Precision at an IoU threshold of 0.50), computed separately per lighting level, per weather condition, and per router branch, and compared across three configurations: (i) baseline (no preprocessing), (ii) uniform SCI applied indiscriminately to every frame, and (iii) the proposed luminance-adaptive router. Relative degradation was additionally reported as percentage retention of clean-condition mAP@50. All evaluations used the Ultralytics evaluation loop for YOLOv8n, YOLOv12n, and RT-DETR, and a torchvision-based evaluation loop for Faster R-CNN, applied consistently across all three preprocessing configurations.

\subsection{Edge Deployment Setup}
Edge feasibility was assessed by deploying all four detectors, the SCI module, and the luminance-adaptive router on an NVIDIA Jetson Nano. Per-stage latency (luminance computation, enhancement, detection) and end-to-end frames-per-second (FPS) were measured for each model on each router branch, using 50 timed runs per configuration following a 10-run warm-up period that was excluded from reported statistics. This protocol isolates the fixed cost of luminance computation and SCI enhancement from the variable cost of the downstream detector backbone.

\section{RESULTS}

\subsection{Baseline Robustness Under Degradation}
Under baseline (no-preprocessing) conditions, mAP@50 declined sharply from clean to severe lighting for all four architectures (Fig. 1), with a pronounced non-linear cliff between dusk and night. RT-DETR was the most robust, retaining 98\% of clean performance at dusk and 37\% at severe darkness, while Faster R-CNN retained only 28\% at severe, the weakest of the four. Under weather degradation, light rain caused negligible loss (90–100\% retention) and Harmattan fog produced graduated degradation (50–70\% retention, RT-DETR strongest), but heavy rain caused a catastrophic collapse to 24–31\% retention across all architectures (Fig. 2), substantially worse than any lighting condition. RT-DETR’s global attention mechanism consistently gave it an advantage under partial occlusion (fog, rain), whereas Faster R-CNN’s two-stage region-proposal pipeline was the most fragile across nearly every degraded condition.

\begin{table*}[t]
\caption{Complete mAP@50 Results. D Columns Show Delta vs Baseline}
\label{tab:map_results}
\begin{center}
\begin{tabular}{llccccccc}
\hline
\textbf{Model} & \textbf{System} & \textbf{Clean} & \textbf{Dusk} & \textbf{Night} & \textbf{Severe} & \textbf{Dusk D} & \textbf{Night D} & \textbf{Severe D} \\
\hline
RT-DETR   & Baseline  & 0.883 & 0.861 & 0.612 & 0.324 & --     & --     & --     \\
          & Plain SCI & 0.883 & 0.722 & 0.679 & 0.387 & -0.139 & +0.067 & +0.063 \\
          & Adaptive  & 0.883 & 0.847 & 0.657 & 0.338 & -0.014 & +0.045 & +0.014 \\
\hline
YOLOv12n  & Baseline  & 0.894 & 0.867 & 0.570 & 0.305 & --     & --     & --     \\
          & Plain SCI & 0.894 & 0.706 & 0.601 & 0.368 & -0.161 & +0.031 & +0.063 \\
          & Adaptive  & 0.894 & 0.829 & 0.577 & 0.294 & -0.038 & +0.007 & -0.011 \\
\hline
YOLOv8n   & Baseline  & 0.881 & 0.855 & 0.590 & 0.268 & --     & --     & --     \\
          & Plain SCI & 0.881 & 0.713 & 0.582 & 0.282 & -0.142 & -0.008 & +0.014 \\
          & Adaptive  & 0.881 & 0.836 & 0.554 & 0.218 & -0.019 & -0.036 & -0.050 \\
\hline
Faster R-CNN & Baseline  & 0.779 & 0.755 & 0.548 & 0.215 & --     & --     & --     \\
             & Plain SCI & 0.779 & 0.523 & 0.442 & 0.208 & -0.232 & -0.106 & -0.007 \\
             & Adaptive  & 0.779 & 0.687 & 0.407 & 0.163 & -0.068 & -0.141 & -0.052 \\
\hline
\end{tabular}
\end{center}
\end{table*}

\subsection{Uniform Enhancement is Harmful at Dusk}
Applying SCI uniformly caused significant mAP@50 degradation at dusk across all four architectures (Table I), ranging from -16.1\% (RT-DETR) to -30.7\% (Faster R-CNN). The same pass was conversely beneficial at night and severe darkness for RT-DETR and YOLOv12n, and consistently harmful for Faster R-CNN except at severe. This confirms enhancement value is condition- and architecture-dependent, not uniformly positive.

\begin{figure}[t]
\centerline{\includegraphics[width=\columnwidth]{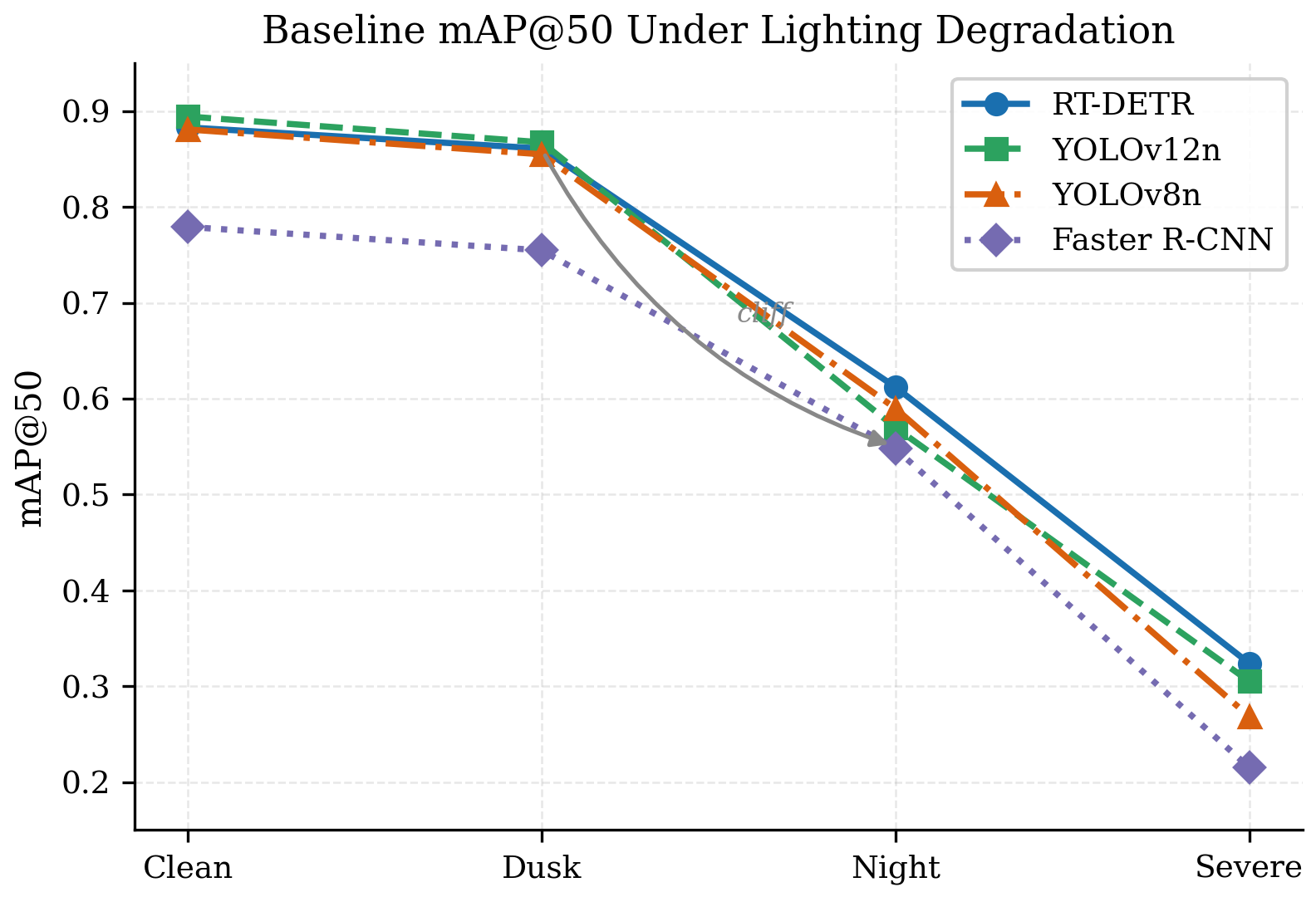}}
\caption{Baseline mAP@50 under lighting degradation.}
\label{fig:baseline_lighting}
\end{figure}

\begin{figure}[t]
\centerline{\includegraphics[width=\columnwidth]{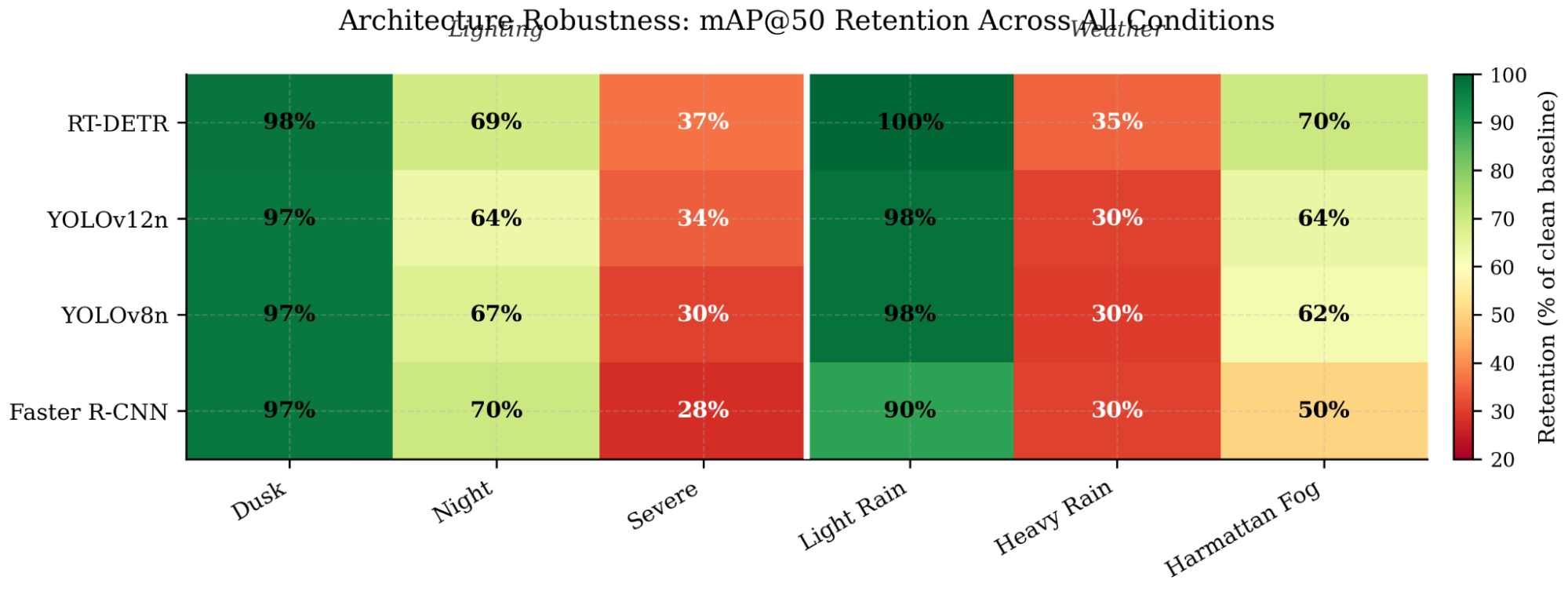}}
\caption{Architecture robustness: mAP@50 retention (percentage of clean baseline) across all lighting and weather conditions.}
\label{fig:architecture-robustness}
\end{figure}

\subsection{Adaptive Router Recovers the Dusk Penalty}
The router substantially recovered the plain-SCI dusk penalty (Table I), reducing the residual loss to 0.001–0.073 mAP@50 below baseline across both validation and held-out test splits, with RT-DETR, YOLOv12n, and YOLOv8n remaining within 0.007 of baseline dusk performance on unseen test data. At night, RT-DETR retained the strongest gain under the router (+0.045 validation, +0.071 test split), YOLOv12n showed consistent recovery (+0.036 test), while YOLOv8n and Faster R-CNN showed additional losses, the latter exceeding its plain-SCI loss and confirming the fixed thresholds are not uniformly optimal for two-stage detectors. Baseline degradation trends were consistent across splits, with absolute mAP@50 values 2–6 points lower on unseen test data as expected.

\begin{table}[t]
\caption{Jetson Nano Pipeline Latency. FPS Thresholds: $\geq$20 real-time, $\geq$15 safety minimum, $<$15 below safety}
\label{tab:latency}
\begin{center}
\small
\setlength{\tabcolsep}{4pt}
\begin{tabular}{llcccc}
\hline
\textbf{Model} & \textbf{Branch} & \textbf{Lum.} & \textbf{Enh.} & \textbf{Det.} & \textbf{FPS} \\
 & & \textbf{(ms)} & \textbf{(ms)} & \textbf{(ms)} & \\
\hline
RT-DETR      & A (Clean)  & 1.2 & 0    & 120.4 & 8.23  \\
             & B (Dusk)   & 1.2 & 8.3  & 121.3 & 7.65  \\
             & C (Severe) & 1.2 & 27.3 & 120.9 & 6.69  \\
\hline
YOLOv12n     & A (Clean)  & 1.2 & 0    & 52.7  & 18.54 \\
             & B (Dusk)   & 1.2 & 13.2 & 52.4  & 14.97 \\
             & C (Severe) & 1.2 & 33.8 & 52.4  & 11.44 \\
\hline
YOLOv8n      & A (Clean)  & 1.2 & 0    & 37.5  & 25.83 \\
             & B (Dusk)   & 1.2 & 16.2 & 37.6  & 18.18 \\
             & C (Severe) & 1.2 & 36.0 & 36.8  & 13.52 \\
\hline
Faster R-CNN & A (Clean)  & 2.0 & 0    & 224.5 & 4.42  \\
             & B (Dusk)   & 2.0 & 8.3  & 223.6 & 4.28  \\
             & C (Severe) & 2.0 & 38.5 & 222.0 & 3.81  \\
\hline
\end{tabular}
\end{center}
\end{table}

\subsection{Edge Deployment Throughput and Latency}
YOLOv8n was the only architecture approaching real-time operation (15-25.8 FPS) across all branches (Table II); YOLOv12n remained close (11.4-18.6 FPS). RT-DETR and Faster R-CNN fell well short at every branch despite stronger degradation robustness, confirming a direct accuracy-latency trade-off. SCI added a fixed overhead independent of detector (16 ms Branch B, 36 ms Branch C), making its relative cost largest for the fastest models.

\section{DISCUSSION}
These results show that front-end enhancement is not a universally safe intervention: SCI improves detection only when the input signal is genuinely inadequate, and actively degrades it when applied to scenes, such as dusk, that already carry sufficient luminance for the detector’s feature extractor. The luminance-adaptive router demonstrates that a simple, training-free, threshold-based decision is sufficient to capture most of this effect, recovering the majority of the dusk penalty without any additional model training. RT-DETR’s consistent robustness across both lighting and weather conditions is plausibly attributable to its global self-attention mechanism, which is less reliant on local high-frequency texture than the convolutional backbones of the YOLO models and Faster R-CNN. 

\section{CONCLUSION}
This paper benchmarked four object-detection architectures for road-anomaly detection under physically grounded lighting and weather degradation and showed that uniform low-light enhancement is actively harmful at dusk, degrading mAP@50 by 16–31\% across architectures. A lightweight, training-free luminance-adaptive router, selecting between no enhancement, moderate SCI, and denoising plus strong SCI based on per-frame Rec. 601 luminance, recovered 71–90\% of the plain-SCI dusk penalty on the validation split and maintained dusk performance within 0.007 of baseline on the held-out test split, with RT-DETR retaining a night-level gain of +0.071 mAP@50 on unseen data. Edge validation on a Jetson Nano confirmed that only YOLOv8n, and marginally YOLOv12n, are suitable for real-time deployment at this hardware tier.

\section{FUTURE WORKS}
Immediate extensions include real-world data collection under Nigerian night-time, Harmattan, and rainy-season conditions to validate the synthetic findings, and compounded degradation benchmarking combining lighting and weather simultaneously. Weather-specific restoration modules (dehazing, deraining), per-architecture threshold calibration for two-stage detectors, and model compression for sub-Jetson-Nano deployment remain open directions.


\begin{thebibliography}{00}
\bibitem{b1} Federal Road Safety Corps (FRSC), ``2024 annual road traffic crash report,'' press briefing by Corps Marshal S. Mohammed, Abuja, Nigeria, Jan. 2025.
\bibitem{b2} World Health Organization, \textit{Global Status Report on Road Safety 2023}. Geneva, Switzerland: WHO, 2023. [Online]. \\Available: \url{https://www.who.int/publications/i/item/9789240086517}
\bibitem{b3} World Bank Global Road Safety Facility, ``The single organizational model for road safety in Nigeria,'' World Bank Group, Washington, DC, USA, 2023. [Online]. Available: \url{https://www.globalroadsafetyfacility.org/sites/default/files/2023-10/Nigeria%20The%20%E2%80%9CSingle%20Organization%20Road%20Safety%E2%80%9D%20Institutional%20Model%2C%20its%20Efficacy%20and%20Replicability.pdf}
\bibitem{b4} G. Jocher, A. Chaurasia, and J. Qiu, ``Ultralytics YOLOv8,'' 2023. [Online]. Available: \url{https://github.com/ultralytics/ultralytics}
\bibitem{b5} Y. Safyari, M. Mahdianpari, and H. Shiri, ``A review of vision-based pothole detection methods using computer vision and machine learning,'' \textit{Sensors}, vol. 24, no. 17, p. 5652, 2024.
\bibitem{b6} A. C. Anuforom, ``Spatial distribution and temporal variability of Harmattan dust haze in sub-Sahel West Africa,'' \textit{Atmos. Environ.}, vol. 41, no. 39, pp. 9079--9090, 2007.
\bibitem{b7} C. Guo \textit{et al.}, ``Zero-reference deep curve estimation for low-light image enhancement,'' in \textit{Proc. IEEE/CVF Conf. Comput. Vis. Pattern Recognit. (CVPR)}, 2020, pp. 1780--1789.
\bibitem{b8} Y. Jiang \textit{et al.}, ``EnlightenGAN: Deep light enhancement without paired supervision,'' \textit{IEEE Trans. Image Process.}, vol. 30, pp. 2340--2349, 2021.
\bibitem{b9} L. Ma, T. Ma, R. Liu, X. Fan, and Z. Luo, ``Toward fast, flexible, and robust low-light image enhancement,'' in \textit{Proc. IEEE/CVF Conf. Comput. Vis. Pattern Recognit. (CVPR)}, 2022, pp. 5637--5646.
\bibitem{b10} H. Liu, F. Jin, H. Zeng, H. Pu, and B. Fan, ``Image enhancement guided object detection in visually degraded scenes,'' \textit{IEEE Trans. Neural Netw. Learn. Syst.}, vol. 35, no. 10, pp. 14164--14177, 2023.
\bibitem{b11} Y. Wang \textit{et al.}, ``Is underwater image enhancement all object detectors need?,'' \textit{IEEE J. Ocean. Eng.}, vol. 49, no. 2, pp. 606--621, 2023.
\bibitem{b12} Y. Tian, Q. Ye, and D. Doermann, ``YOLOv12: Attention-centric real-time object detectors,'' in \textit{Proc. Adv. Neural Inf. Process. Syst. (NeurIPS)}, vol. 38, 2026, pp. 78433--78457.
\bibitem{b13} Y. Zhao \textit{et al.}, ``DETRs beat YOLOs on real-time object detection,'' in \textit{Proc. IEEE/CVF Conf. Comput. Vis. Pattern Recognit. (CVPR)}, 2024, pp. 16965--16974.
\bibitem{b14} S. Ren, K. He, R. Girshick, and J. Sun, ``Faster R-CNN: Towards real-time object detection with region proposal networks,'' in \textit{Proc. Adv. Neural Inf. Process. Syst. (NIPS)}, vol. 28, 2015.
\bibitem{b15} A. Dhiman and R. Klette, ``Pothole detection using computer vision and learning,'' \textit{IEEE Trans. Intell. Transp. Syst.}, vol. 21, no. 8, pp. 3536--3550, 2019.
\bibitem{b16} D. Arya \textit{et al.}, ``Crowdsensing-based road damage detection challenge (CRDDC'2022),'' in \textit{Proc. IEEE Int. Conf. Big Data (Big Data)}, 2022, pp. 6378--6386.
\bibitem{b17} D. Arya, H. Maeda, S. K. Ghosh, D. Toshniwal, and Y. Sekimoto, ``RDD2022: A multi-national image dataset for automatic road damage detection,'' \textit{Geosci. Data J.}, vol. 11, no. 4, pp. 846--862, 2024.
\bibitem{b18} Y. Zanevych \textit{et al.}, ``Evaluation of pothole detection performance using deep learning models under low-light conditions,'' \textit{Sustainability}, vol. 16, no. 24, p. 10964, 2024.
\bibitem{b19} J. Terven, D.-M. C\'ordova-Esparza, and J.-A. Romero-Gonz\'alez, ``A comprehensive review of YOLO architectures in computer vision: From YOLOv1 to YOLOv8 and YOLO-NAS,'' \textit{Mach. Learn. Knowl. Extr.}, vol. 5, no. 4, pp. 1680--1716, 2023.
\bibitem{b20} K. He, J. Sun, and X. Tang, ``Single image haze removal using dark channel prior,'' \textit{IEEE Trans. Pattern Anal. Mach. Intell.}, vol. 33, no. 12, pp. 2341--2353, 2010.
\bibitem{b21} Z. Chen \textit{et al.}, ``Object detection for autonomous vehicles under adverse weather conditions,'' \textit{Expert Syst. Appl.}, 2025, Art. no. 128994.
\bibitem{b22} S. S. Halder, J.-F. Lalonde, and R. de Charette, ``Physics-based rendering for improving robustness to rain,'' in \textit{Proc. IEEE/CVF Int. Conf. Comput. Vis. (ICCV)}, 2019, pp. 10203--10212.
\bibitem{b23} W. Liu \textit{et al.}, ``Image-adaptive YOLO for object detection in adverse weather conditions,'' in \textit{Proc. AAAI Conf. Artif. Intell.}, vol. 36, no. 2, 2022, pp. 1792--1800.
\bibitem{b24} S. Kalwar \textit{et al.}, ``GDIP: Gated differentiable image processing for object detection in adverse conditions,'' \textit{arXiv preprint arXiv:2209.14922}, 2022.
\bibitem{b25} Q. Qin, K. Chang, M. Huang, and G. Li, ``DENet: Detection-driven enhancement network for object detection under adverse weather conditions,'' in \textit{Proc. Asian Conf. Comput. Vis. (ACCV)}, 2022, pp. 2813--2829.
\bibitem{b26} R. Gao, M. H. Omar, and M. Mahmuddin, ``YOLO-DER: A dynamic enhancement routing framework for adverse weather vehicle detection,'' \textit{Electronics}, vol. 14, no. 24, p. 4851, 2025.
\bibitem{b27} Z. Wei, Y. Wang, L. Sun, A. V. Vasilakos, and L. Wang, ``ClassLIE: Structure- and illumination-adaptive classification for low-light image enhancement,'' \textit{IEEE Trans. Artif. Intell.}, vol. 5, no. 9, pp. 4765--4775, 2024.
\bibitem{b28} H. Feng, G. Mu, S. Zhong, P. Zhang, and T. Yuan, ``Benchmark analysis of YOLO performance on edge intelligence devices,'' \textit{Cryptography}, vol. 6, no. 2, p. 16, 2022.
\bibitem{b29} D. K. Alqahtani, M. A. Cheema, and A. N. Toosi, ``Benchmarking deep learning models for object detection on edge computing devices,'' in \textit{Proc. Int. Conf. Service-Oriented Comput.}, Singapore: Springer, 2024, pp. 142--150.
\bibitem{b30} Rajdalsaniya, ``Pothole detection dataset,'' Kaggle, 2022. [Online]. Available: \url{https://www.kaggle.com/datasets/rajdalsaniya/pothole-detection-dataset}
\bibitem{b31} ``Speed-unmarked-bumb dataset, v3,'' Roboflow Universe, 2023. [Online]. Available: \url{https://universe.roboflow.com/pothole-detection-1nczj/speed-unmarked-bumb}
\end{thebibliography}
\end{document}